\documentclass[runningheads]{llncs}
\usepackage[T1]{fontenc}
\usepackage{amsmath}
\usepackage{amssymb}
\usepackage{graphicx,verbatim}
\usepackage{utfsym}
\usepackage{esvect}

\usepackage{multirow}
\usepackage{booktabs}
\usepackage{color}
\usepackage{hyperref} 
\hypersetup{colorlinks=true, linkcolor=blue, citecolor=blue, urlcolor=blue} 
\usepackage[misc]{ifsym} 
\usepackage{wrapfig}

\begin{document}
\title{Improved Baselines with Synchronized Encoding for Universal Medical Image Segmentation}
\titlerunning{SyncSAM: Improved Baselines for Universal Medical Image Segmentation}
\author{Sihan Yang\inst{1} \and
Jiadong Feng\inst{1}\textsuperscript{*} \and
Xuande Mi\inst{1}\textsuperscript{*} \and
Haixia Bi \inst{2}\textsuperscript{(\Letter)} 
\and
\\
Hai Zhang \inst{3,5} \and
Jian Sun \inst{4,5}
}
%index{Yang, Sihan}
%index{Feng, Jiadong}
%index{Mi, Xuande}
%index{Bi, Haixia}
%index{Zhang, Hai}
%index{Sun, Jian}
%
\authorrunning{S. Yang. et al.}
\institute{Xi’an Jiaotong University, Xi’an, China
\and
School of Information and Communications Engineering, Xi’an Jiaotong University, Xi'an, China
\\
\email{haixia.bi@xjtu.edu.cn}
\and
School of Mathematics, Northwest University, Xi'an, China
\and
School of Mathematics and Statistics, Xi’an Jiaotong University, Xi'an, China
\and
Pazhou Laboratory (Huangpu), Guangzhou, China
}
\maketitle              % typeset the header of the contribution
\footnotetext{\textsuperscript{*} Equal contributions.}
\begin{abstract}
Large foundation models, known for their strong zero-shot generalization capabilities, can be applied to a wide range of downstream tasks. However, developing foundation models for medical image segmentation poses a significant challenge due to the domain gap between natural and medical images. While fine-tuning techniques based on the Segment Anything Model (SAM) have been explored, they primarily focus on scaling up data or refining inference strategies without incorporating domain-specific architectural designs, limiting their zero-shot performance. To optimize segmentation performance under standard inference settings and provide a strong baseline for future research, we introduce SyncSAM, which employs a synchronized dual-branch encoder that integrates convolution and Transformer features in a synchronized manner to enhance medical image encoding, and a multi-scale dual-branch decoder to preserve image details. SyncSAM is trained on two of the largest medical image segmentation datasets, SA-Med2D-20M and IMed-361M, resulting in a series of pre-trained models for universal medical image segmentation. Experimental results demonstrate that SyncSAM not only achieves state-of-the-art performance on test sets but also exhibits strong zero-shot capabilities on unseen datasets. Code and checkpoints are available at \url{https://github.com/Hhankyangg/SyncSAM}.

\end{abstract}
\section{Introduction}

Large foundation models pre-trained on extensive datasets provide strong zero-shot capabilities, making them effective for diverse downstream tasks~\cite{an2025unictokens,luo2024llm,an2025concept,bi2023boosting}. A key example is the Segment Anything Model (SAM)~\cite{kirillov2023segment}, an interactive segmentation model trained on large datasets. Leveraging its robust zero-shot performance, SAM has not only established a stable foundation for segmentation tasks~\cite{ren2024grounded} but also advanced various visual and multimodal applications~\cite{lin2025perceive,an2024mc}.

Medical image segmentation is a fundamental task in medical imaging analysis, playing a crucial role in disease diagnosis, treatment planning, and disease progression monitoring~\cite{de2018clinically}. With the emergence of SAM, a corpus of researchers have explored its potential in  medical image segmentation. These works can be categorized based on the dataset scale: leveraging SAM for specialist models on small datasets and for foundation models on large-scale datasets. For specialist models trained on small datasets, approaches include Parameter-Efficient Fine-Tuning (PEFT)~\cite{wu2025medical} and architectural modifications such as hierarchical decoders~\cite{cheng2024unleashing} and prompt-decoupled mask decoders~\cite{gao2024desam}. While effective on small datasets, their performance on large datasets remains unverified. For foundation models trained on large-scale datasets, two primary fine-tuning strategies exist. The first fine-tunes SAM at different parameter scales using either PEFT techniques~\cite{cheng2023sam} or full-parameter fine-tuning~\cite{ma2024segment}, yet these models often lack domain-specific architectural enhancements, leading to suboptimal performance. The second modifies the inference process~\cite{zhu2024medical,wu2024one} to improve segmentation accuracy, incorporating advanced interactive methods or support-set-based strategies. While effective in complex prompt settings, these modifications may limit broader applicability to downstream tasks~\cite{koleilat2024medclip}.  
 
To fill the gap of foundation models in integrating expert-designed modules, we introduce SyncSAM in this work. Inspired by the advantages of Convolutional Neural Network (CNN) in capturing fine-grained features and dealing with noisy images, we incorporate a CNN branch into SAM’s image encoder, injecting medical-specific domain bias into the Vision Transformer(ViT)~\cite{dosovitskiy2020image} backbone. Moreover, we propose a synchronized fusion strategy that performs stage-wise fusion of ViT and CNN features rather than merging them in a single step. This progressive alignment of local and global representations enhances contextual understanding of medical images. To further improve segmentation, we devise a multi-scale dual-branch decoder that leverages early-stage encoder features to refine fine-grained edge details. By combining the synchronized dual-branch encoder and the multi-scale dual-branch decoder, SyncSAM effectively enhances feature representation, enabling more precise segmentation for medical images.

We train four versions of SyncSAM on two of the largest 2D medical image segmentation datasets, SA-Med2D-20M~\cite{ye2023sa} and IMed-361M~\cite{cheng2025interactive}. Experimental results on test sets confirm state-of-the-art (SOTA) performance, while zero-shot evaluations on six unseen datasets demonstrate that SyncSAM outperforms all existing medical foundation models using simple box prompts.

This work makes three key contributions.  
(1) We introduce SyncSAM, a novel foundation model tailored for medical image segmentation, filling the gap in large-scale SAM-based foundation models with expert-designed modules.  
(2) We train and evaluate SyncSAM on two of the largest 2D medical segmentation datasets, achieving SOTA performance on test sets while demonstrating strong scalability across different dataset sizes.  
(3) We extensively conduct zero-shot segmentation experiments on six external datasets, showing that SyncSAM surpasses existing medical foundation models under simple bounding box prompts, highlighting its strong potential for wide downstream applications.

\begin{figure}[t]
\centering
\includegraphics[width=0.9\textwidth]{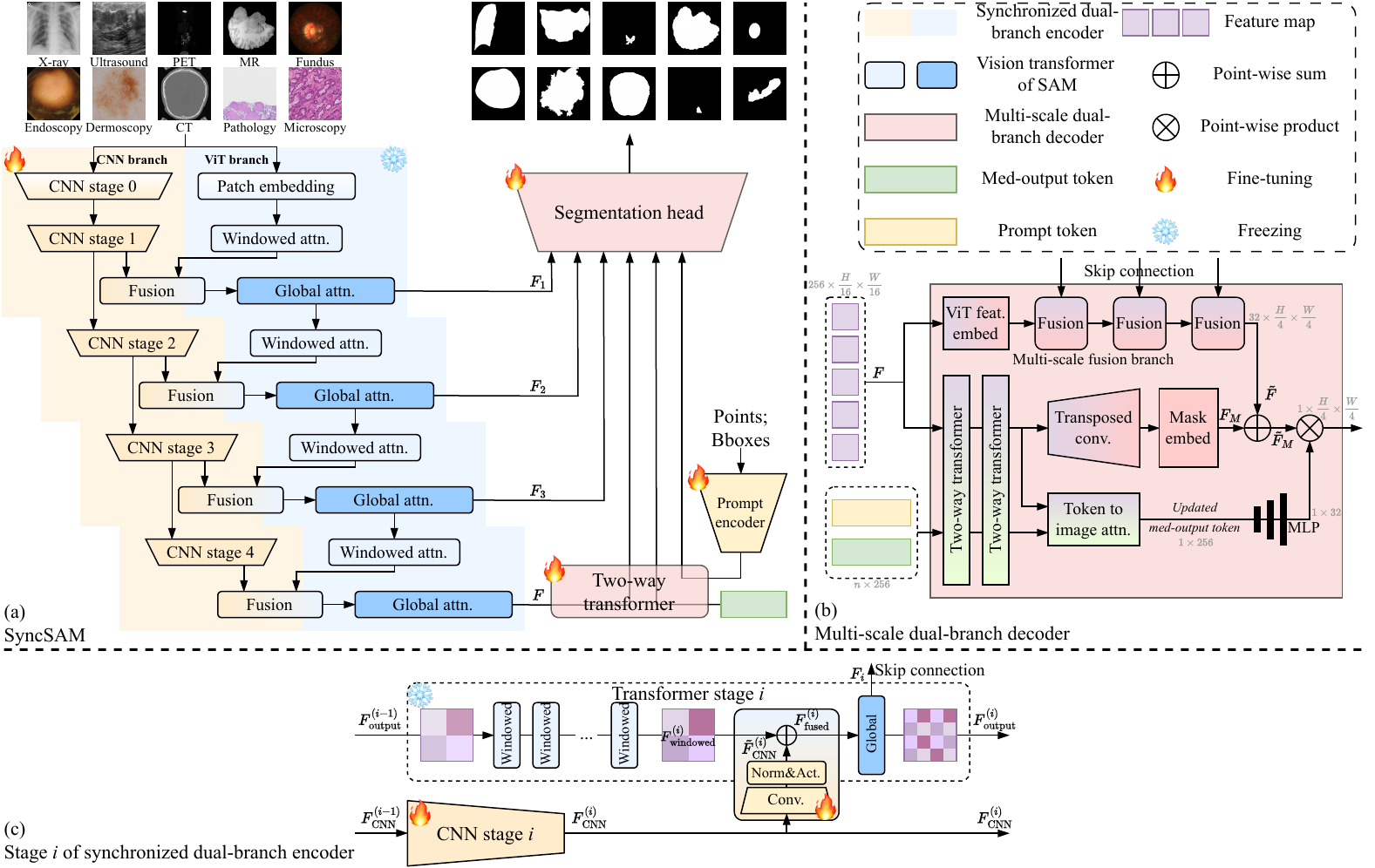}
\caption{(a) Overview of SyncSAM. (b) Multi-scale dual-branch decoder. (c) Synchronized fusion in the synchronized dual-branch encoder at the $i$-th stage.}
\label{model_overview}
\end{figure}

\section{Method}

\subsection{Overview}

Figure.~\ref{model_overview}(a) illustrates the framework of our proposed SyncSAM, which aims to enhance medical image segmentation with a synchronized dual-branch architecture. Given a pre-processed medical image $I \in \mathbb{R}^{3 \times H \times W}$ and a prompt $c$ (points, boxes, or masks), SyncSAM predicts the corresponding binary mask $M^{(c)}$:

\begin{equation}
M^{(c)} = {\text{SyncSAM}}(I, c)
\end{equation}

SyncSAM consists of three main components: a synchronized dual-branch encoder, a prompt encoder, and a multi-scale dual-branch decoder. The synchronized dual-branch encoder retains SAM’s ViT while integrating a CNN-based synchronized fusion branch to align features at each stage [shown in Fig.~\ref{model_overview}(b)]. The prompt encoder, inherited from SAM, transforms sparse and dense prompts into vector representations, referred to as prompt token. The decoder then integrates feature maps with prompt token, while the multi-scale fusion branch [detailed in Fig.~\ref{model_overview}(c)] enhances feature refinement. Together with the proposed Med-Output Token, these components collaboratively produce the final segmentation output. The following sections detail each component.

\subsection{Synchronized Dual-Branch Encoder}

The synchronized dual-branch encoder serves as the image encoder in the model, encoding a preprocessed image $I \in \mathbb{R}^{3 \times H \times W}$ into a rich semantic feature map $F \in \mathbb{R}^{256 \times \frac{H}{16} \times \frac{W}{16} }$. To be specific, the encoder consists of two parallel branches, i.e., a ViT branch and a CNN branch. The ViT branch, inherited from SAM, captures long-range dependencies and global context while remaining frozen to limit the number of trainable parameters. In contrast, the CNN branch extracts local fine-grained features, which are particularly advantageous in handling noise and structural variations commonly found in medical images. Initially, the input image undergoes patch embedding in the ViT branch and stage 0 processing in the CNN branch as a preprocessing step. The image is then progressively processed through four consecutive stages in both branches.

To effectively integrate features from both branches, we design a synchronized fusion mechanism that merges encoded representations at each stage. At stage $i$, the ViT branch leverages windowed attention transformer blocks to refine the output feature map $F_{\text{output}}^{(i-1)}$ from the previous stage, producing $F_{\text{windowed}}^{(i)}$. Simultaneously, the CNN branch extracts local and domain-specific features from $F_{\text{CNN}}^{(i-1)}$, generating $F_{\text{CNN}}^{(i)}$. To ensure compatibility between the two feature maps, $F_{\text{CNN}}^{(i)}$ undergoes convolution to match the ViT feature dimensions, yielding $\tilde{F}_{\text{CNN}}^{(i)}$. This transformed CNN feature map is then point-wise added to $F_{\text{windowed}}^{(i)}$, forming the fused feature map $F_{\text{fused}}^{(i)}$. The fused representation is further refined through a global attention transformer block, producing the stage output $F_{\text{output}}^{(i)}$. Going through all four stages, the encoded feature map is further refined by the ViT neck layer, generating the final output $F$.

\subsection{Multi-Scale Dual-Branch Decoder}

The multi-scale dual-branch decoder introduces two key modifications compared to SAM’s mask decoder, as shown in Fig.~\ref{model_overview}(b). 

\textbf{Med-Output Token.}  
Inspired by SAM-HQ~\cite{ke2023segment}, we replace SAM’s IoU prediction head and multiple mask tokens with a single trainable Med-Output Token ($1 \times 256$). In SAM, these tokens resolve ambiguities in point-based prompts for natural images. However, such ambiguities are rare in medical image segmentation, making multiple masks and IoU prediction unnecessary.

\textbf{Multi-Scale Fusion Branch.}  
A multi-scale fusion branch is incorporated into the mask decoder, as early-stage features preserve more fine-grained edge details~\cite{ronneberger2015u}. This branch fuses feature maps $F_i$ from the first three ViT stages with a dimensionally adjusted $F$ from the image encoder, generating a multi-scale feature map $\tilde{F} \in \mathbb{R}^{32 \times \frac{H}{4} \times \frac{W}{4}}$. The feature fusion is performed via a series of convolutional layers. $\tilde{F}$ is then point-wise added to the mask feature $F_M$, which is derived from $F$ using a Two-Way Transformer to integrate information from the output tokens, including prompt tokens and the Med-Output Token.

\section{Experiments}

\subsection{Training of SyncSAM}

\subsubsection{Datasets.}

We train our model on two of the largest 2D medical image segmentation datasets to date: SA-Med2D-20M~\cite{ye2023sa} and IMed-361M~\cite{cheng2025interactive}. This resulted in two versions of SyncSAM, SyncSAM-SAMed and SyncSAM-IMed, to demonstrate the model's scalability across different datasets. Specifically, SA-Med2D-20M contains 3.6 million images and 15.8 million ground truth masks, while IMed-361M includes 2.2 million images and 62.6 million ground truth masks, with modality distributions as shown in the Tab.~\ref{tab:modal_stat}. For SA-Med2D-20M, which does not provide a predefined training-test split, we randomly split the data with 80\% used for training and the remaining 20\% for testing. For IMed-361M, we follow the official dataset split.

\begin{table}[t]
\centering
\scriptsize 
\setlength{\tabcolsep}{3mm}
\renewcommand{\arraystretch}{1.0}  
\caption{The modality distributions in SA-Med2D-20M and IMed-361M.}

\begin{tabular}{l|c|c|c|c}
\hline
\textbf{Modality} & \textbf{CT} & \textbf{Endoscopy} & \textbf{PET} & \textbf{Fundus} \\
\hline
\textbf{Images in SA-Med2D-20M}   & 1,645,894 & 4,290  & 5,410  & 1,445  \\
\textbf{Masks in SA-Med2D-20M}    & 5,533,808 & 15,469 & 6,282  & 1,741  \\
\hline
\textbf{Images in IMed-361M}      & 1,726,089 & 52,568 & 0      & 1,275  \\
\textbf{Masks in IMed-361M}       & 61,190,317 & 52,568 & 0     & 2,348  \\
\hline
\hline
\textbf{Modality} & \textbf{MR} & \textbf{Dermoscopy} & \textbf{X-ray} & \textbf{Ultrasound} \\
\hline
\textbf{Images in SA-Med2D-20M}   & 1,967,254 & 6,698  & 5,581  & 2,590  \\
\textbf{Masks in SA-Med2D-20M}    & 5,533,808 & 6,858  & 7,067  & 2,590  \\
\hline
\textbf{Images in IMed-361M}      & 410,261   & 6,123  & 566    & 45,103 \\
\textbf{Masks in IMed-361M}       & 1,291,195 & 6,123  & 566    & 45,103 \\
\hline
\end{tabular}
\label{tab:modal_stat}
\end{table}

\subsubsection{Loss Function.}

We modify SAM’s loss function by removing the IoU prediction loss, as the IoU head is no longer used. The final loss function is defined as:

\begin{equation}
    \mathcal{L} = \lambda \mathcal{L}_{Dice} + \mathcal{L}_{Focal}
\end{equation} 

\subsubsection{Training Details.}

Following previous studies~\cite{kirillov2023segment,ma2024segment,cheng2023sam}, we train SyncSAM using interactive segmentation simulation, and randomly select five masks per image for a training step.  We empirically set $\lambda = 20$. For preprocessing, images are padded to square shapes and resized to $256 \times 256$ before being fed into the model. We use the ViT-B version of SAM, keeping its image encoder frozen while integrating different CNN branches—ResNet50 and ResNet34~\cite{he2016deep}. The corresponding models, SyncSAM-50 and SyncSAM-34, contain approximately 138 million and 116 million parameters, respectively, of which about 48 million and 26 million are trainable. Training is conducted for 12 epochs with an initial learning rate of 0.0001. We use 8 NVIDIA Tesla A100 GPUs, processing 50 images and 250 masks per GPU per step.

\subsection{Experimental Setup}

We assess SyncSAM through three key experiments: 
(1) test set evaluations against models trained on the same dataset (Sec.~\ref{sec:testset}), 
(2) zero-shot segmentation performance on external datasets (Sec.~\ref{sec:zeroshot}), and 
(3) ablation studies analyzing the impact of individual components (Sec.~\ref{sec:ablation}). 
In the following tables, ``SAM'' denotes direct inference with SAM, while ``FT-SAM'' refers to SAM with only the mask decoder trained. All results are reported as Dice Similarity Coefficient (DSC) percentages, with \textbf{best} and \underline{second best} performances highlighted. All reported scores are averaged over three independent runs.

\subsection{Test Set Performance Comparison}
\label{sec:testset}

\begin{table*}[t]
\centering
\scriptsize
\setlength{\tabcolsep}{1.2mm}  % Adjust column spacing
\renewcommand{\arraystretch}{1.0}  % Adjust row height

\begin{minipage}{0.49\linewidth}
    \centering
    \caption{Performance comparison on the test set of SA-Med2D-20M.}
    \begin{tabular}{l||ccc}
    \toprule
    \multirow{2}{*}{\textbf{Model}} & \multicolumn{3}{c}{\textbf{DSC}} \\ 
     & Bbox & 1 pt & 5 pts \\ 
    \midrule
    SAM & 66.6 & 24.5 & 52.8\\ 
    \midrule
    MedSAM  & 80.8 & \usym{2717}& \usym{2717}\\
    SAM-Med2D  & 78.2 & 68.3 & 76.7\\
    FT-SAM & 74.6 & 61.1 & 73.2\\
    \midrule
    \textbf{SyncSAM-SAMed-34}  & \underline{87.0} & \underline{74.1} & \underline{87.6}\\
    \textbf{SyncSAM-SAMed-50} & \textbf{88.2} &\textbf{74.8} & \textbf{88.5}\\
    \bottomrule
    \end{tabular}
    \label{tab:test-samed20m}
\end{minipage}
\hfill
\begin{minipage}{0.49\linewidth}
    \centering
    \caption{Performance comparison on the test set of IMed-361M.}
    \begin{tabular}{l||ccc}
    \toprule
    \multirow{2}{*}{\textbf{Model}} & \multicolumn{3}{c}{\textbf{DSC}} \\ 
     & Bbox & 1 pt & 5 pts \\ 
    \midrule
    SAM & 67.2 & 23.3 &51.4\\ 
    \midrule
    MedSAM  & 83.8 & \usym{2717}& \usym{2717}\\
    SAM-Med2D  & 82.3 & 78.7 & 83.6\\
    FT-SAM & 80.1 & 78.3 &81.9\\
    \midrule
    \textbf{SyncSAM-IMed-34}  & \underline{88.2} & \underline{86.9} & \underline{89.1}\\
    \textbf{SyncSAM-IMed-50} & \textbf{89.6} & \textbf{87.3} & \textbf{89.7}\\
    \bottomrule
    \end{tabular}
    \label{tab:test-imed-361m}
\end{minipage}
\end{table*}

In the test set comparisons, to ensure fairness, all models except SAM~\cite{kirillov2023segment} (which performs direct inference) are trained on the same dataset. The models—MedSAM~\cite{ma2024segment}, SAM-Med2D~\cite{cheng2023sam}, and FT-SAM—correspond to fully fine-tuned SAM, adapter-tuned SAM, and SAM with only the mask decoder trained, respectively. For interaction modes, we tested three types of prompts: bounding box prompt, a random point prompt, and interactive sequential point inputs (5 points). However, since MedSAM does not implement point-based interaction, it was not evaluated for this mode. The results in Tab.~\ref{tab:test-samed20m} and Tab.~\ref{tab:test-imed-361m} correspond to two different test sets, demonstrating that our two versions of SyncSAM outperform all other methods across both training datasets. For MedSAM, SAM-Med2D, and FT-SAM, as the number of trainable parameters decreases, the model performance also drops to varying degrees. However, compared to the fully fine-tuned MedSAM, our model achieves higher DSC scores with less than half the number of trainable parameters, thanks to SyncSAM's model design.

\subsection{Zero-Shot Segmentation Performance Comparison}
\label{sec:zeroshot}

We conduct zero-shot segmentation experiments on six unseen datasets: STS~\cite{wang2024sts}, HNTSMRG~\cite{wahid_2024_11199559}, CURVAS~\cite{riera_marin_2024_12687192}, COSAS24~\cite{cosas24}, KPIs~\cite{tang2024holohisto}, and EBHI~\cite{ebhi}, which have not been used in the training sets of any of the compared models, covering five modalities in total. The compared models, besides SAM~\cite{kirillov2023segment}, include recent SOTA foundation medical image segmentation models: SAM-Med2D~\cite{cheng2023sam}, MedSAM~\cite{ma2024segment}, the ScribblePrompt~\cite{wong2024scribbleprompt} series, and IMIS-Net~\cite{cheng2025interactive}. To directly evaluate the zero-shot segmentation performance of each base model, we use the widely adopted bounding box prompt and perform direct inference with each model. The results are shown in Tab.~\ref{tab:zeroshot}.

\begin{table}[t]
\centering
\footnotesize
\setlength{\tabcolsep}{0.3mm}  % Adjust column spacing
\renewcommand{\arraystretch}{1.1}  % Adjust row height
\caption{Zero-shot segmentation performance on 6 unseen datasets.}
\begin{tabular}{l||ccc|ccc}
\toprule
\multirow{2}{*}{\textbf{Model}} & \textbf{MR} & {\textbf{CT}} & \textbf{X-ray} & \textbf{Microscopy} & \multicolumn{2}{c}{\textbf{Pathology}} \\ 
& \textbf{HNTSMRG}  & \textbf{CURVAS} & \textbf{STS} & \textbf{COSAS24} & \textbf{KPIs} & \textbf{EBHI} \\ 
\midrule
SAM         & 70.2 & 82.1 & 63.4 & \textbf{66.2} & 84.8 & \textbf{80.6} \\ 
\midrule
 SAM-Med2D   & 83.7  & 91.5 & 70.1 & 61.9 & 82.8 & 71.6 \\
MedSAM      & 62.3  & 65.2 & 59.0 & 59.7 & 70.4 & 77.0 \\ 
 ScribblePrompt-UNet & 61.2 & 61.0 & 48.1 & 45.1 & 49.1 & 46.7 \\
ScribblePrompt-SAM & 54.2 & 60.2 & 49.4 & 43.4 & 50.6 & 48.3 \\
 IMIS-Net & 80.6 & 81.7 & 64.6 & 51.6 & 77.4 & 70.5 \\
\midrule
\textbf{SyncSAM-SAMed-34} & 83.9 & 92.4 & \textbf{73.2} & \underline{65.9} &  85.8 & \underline{80.0} \\
 \textbf{SyncSAM-SAMed-50} & 84.8 & 93.0 & \underline{72.9} & 64.5 & \textbf{86.7} & 79.2 \\
\textbf{SyncSAM-IMed-34} & \underline{85.6} & \underline{94.2} & 68.3 & 64.8 & 84.9 & 78.1 \\
 \textbf{SyncSAM-IMed-50} & \textbf{87.2} & \textbf{94.7} & 68.5 & 62.7 & \underline{85.9} & 78.9 \\
\bottomrule
\end{tabular}
\label{tab:zeroshot}
\end{table}

\begin{figure}[!ht]
\centering
\includegraphics[width=0.9\textwidth]{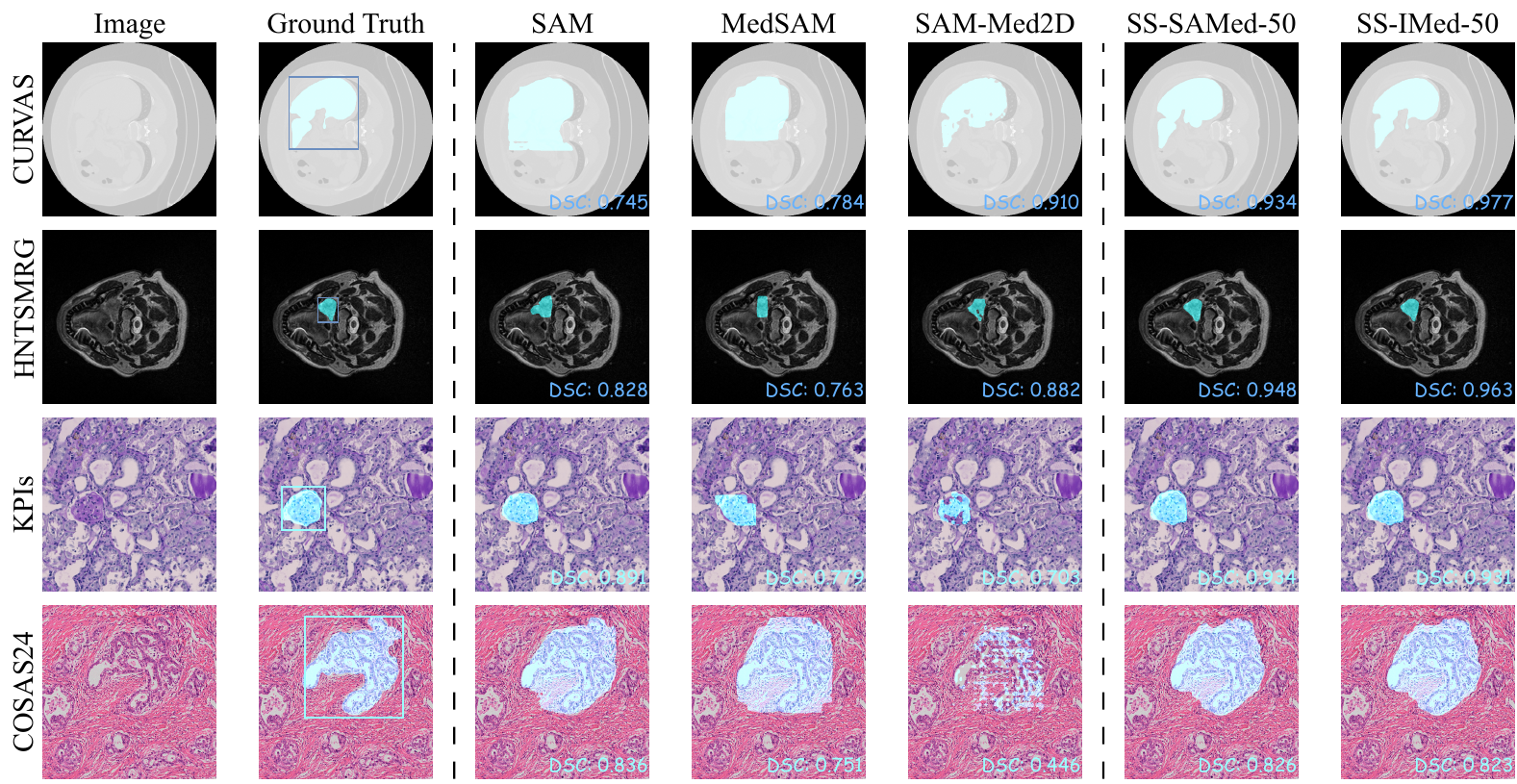}
\caption{Example predictions of zero-shot segmentation. SS = SyncSAM.}
\label{zeroshotshow}
\end{figure}

For MedSAM, due to its limited dataset size, its generalization to medical images remains flawed. As for the ScribblePrompt series, which focuses on training with various interaction methods, its performance under bounding box prompts is unstable. Next, we discuss the experimental results by dataset. 
In common large-scale datasets, CT and MR modalities typically make up about 90\% of the dataset, while other modalities like X-ray, Endoscopy,  Ultrasound, et al. usually account for 0.1\% to 10\%. Microscopy and Pathology data are usually below 0.1\%, and these two modalities are even absent from the released SA-Med2D-20M and IMed-361M datasets. Therefore, we will present the results in three parts based on modality distribution.  
(1) For MR and CT data, the large volume of training samples enables most models to generalize well, with SyncSAM models surpassing all other comparisons. Among the SyncSAM variants, the model trained on IMed-361M performs better due to its larger amount of modality-specific training data.  
(2) For the STS dataset, where X-ray is underrepresented, most models exhibit suboptimal performance. However, SyncSAM-SAMed-34 still outperforms SAM by nearly 10\%. In contrast, SyncSAM trained on IMed-361M, which contains very few X-ray samples, experiences a notable performance drop.  
(3) For Microscopy and Pathology data, where training images are extremely scarce or entirely absent, most models perform worse than SAM due to the reduced generalization ability after fine-tuning on medical datasets. However, SyncSAM benefits from its fully frozen ViT branch, which helps preserve SAM’s encoding ability for diverse images. As a result, SyncSAM maintains a smaller performance gap compared to SAM on these datasets and in some cases even surpasses SAM.

\subsection{Ablation study}
\label{sec:ablation}

We conduct ablation experiments on SA-Med2D-20M~\cite{ye2023sa}, using bounding boxes as prompts to evaluate the impact of each module in SyncSAM, as shown in Tab.~\ref{tab:ablation}. 
Overall, fully fine-tuned SAM achieves a DSC of 80.8 (row 3), while our synchronized encoding variant reaches 85.5 (row 4), highlighting the advantage of our architectural design. Specifically, rows 1 and 4 demonstrate that incorporating the synchronized dual-branch encoder alone improves DSC by 18.9\%. Removing the ViT branch (rows 5, 7, and 8) leads to inferior performance compared to the dual-branch configuration, confirming the necessity of retaining ViT for contextual representation.
Rows 11 and 12 compare different fusion strategies, showing that synchronized fusion improves DSC by 3.3\%, validating its effectiveness. Furthermore, we assess the impact of decoder modifications across various settings (rows 4 vs. 9 vs. 12, rows 5 vs. 7 vs. 8, and rows 2 vs. 6), all of which demonstrate consistent performance gains from decoder enhancements.
Lastly, comparing row 10 and row 12 shows that replacing ResNet-34 with ResNet-50 improves DSC by 1.2\%, suggesting that our model benefits from increased capacity and is scalable with larger CNN backbones.

\begin{table*}[t]
\centering
\scriptsize
\setlength{\tabcolsep}{0.8mm}
\renewcommand{\arraystretch}{1.1}
\caption{Ablation. $\triangle$ = single-step fusion; $\bigstar$ = synchronized fusion; RN = ResNet.}
\begin{tabular}{c|l||cc|cccc|c}
\toprule
\textbf{\#} & \multirow{2}{*}{\textbf{Model}} & \multicolumn{2}{c|}{\textbf{Decoder trained with}} & \multicolumn{4}{c|}{\textbf{Image Encoder}} & \textbf{DSC} \\ 
& & \textbf{Multi-scale} & \textbf{Med-token} & \textbf{ViT (fixed)} & \textbf{RN-34} & \textbf{RN-50} & \textbf{Fusion} & \textbf{Test set} \\ 
\midrule
1 & \textbf{SAM (fixed)} &  &  & $\checkmark$   &    &  & & 66.6 \\
2 & \textbf{FT-SAM} &  &  & $\checkmark$   &    &  & & 74.6 \\ 
3 & \textbf{SAM} &  &  &    &    &  & & 80.8 \\
\midrule
4 &  &  &  & $\checkmark$   &    &  $\checkmark$ &$\bigstar$& 85.5 \\
5 & &  &  &  & & $\checkmark$ & & 79.1 \\
6 & &  $\checkmark$ & $\checkmark$ & $\checkmark$  &   &  &  & 78.2 \\
7 & &  $\checkmark$ & & &  &  $\checkmark$   & & 82.7 \\
8 &  \textbf{SyncSAM} &  $\checkmark$ & $\checkmark$ &  &    &  $\checkmark$   & & 84.6 \\
9 & &  & $\checkmark$ & $\checkmark$ &    &  $\checkmark$   & $\bigstar$& 86.6 \\
10 & &  $\checkmark$ & $\checkmark$ & $\checkmark$ &  $\checkmark$  &  &$\bigstar$&  \underline{87.0} \\
11 & &  $\checkmark$ & $\checkmark$ & $\checkmark$ &    &  $\checkmark$   &$\triangle$ & 84.9 \\
12 & &  $\checkmark$ & $\checkmark$ & $\checkmark$ &    &  $\checkmark$   & $\bigstar$ &  \textbf{88.2} \\
\bottomrule
\end{tabular}
\label{tab:ablation}
\end{table*}

\section{Conclusion}

We introduce SyncSAM, a novel foundation model that combines the SAM with a synchronized dual-branch encoder and a multi-scale dual-branch decoder tailored for medical image segmentation. Trained on the two largest datasets, SA-Med2D-20M and IMed-361M, SyncSAM achieves SOTA performance and demonstrates strong zero-shot generalization on unseen datasets, making it a powerful baseline for medical image segmentation.
In the future, we plan to probe into multi-modality foundation models for medical images.

\begin{credits}
\subsubsection{\ackname} This work was supported by NSFC under Grant 12326615, 42201394 and 12426313, and Key R\&D Program of Shaanxi Province under Grant 2025CY-YBXM-040.

\subsubsection{\discintname}
The authors have no competing interests to declare that are relevant to the content of this article.

% \subsubsection{\discintname}
% It is now necessary to declare any competing interests or to specifically
% state that the authors have no competing interests. Please place the
% statement with a bold run-in heading in small font size beneath the
% (optional) acknowledgments\footnote{If EquinOCS, our proceedings submission
% system, is used, then the disclaimer can be provided directly in the system.},
% for example: The authors have no competing interests to declare that are relevant to the content of this article.
% Or: Author A has received research
% grants from Company W. Author B has received a speaker honorarium from
% Company X and owns stock in Company Y. Author C is a member of committee Z.
\end{credits}

\bibliographystyle{splncs04}
\bibliography{Paper-1708}

\end{document}